\documentclass[letterpaper, 10 pt, conference]{ieeeconf}  % Comment this line out if you need a4paper

\IEEEoverridecommandlockouts                              % This command is only needed if 
\usepackage{graphicx}
\usepackage{amssymb}
\usepackage{booktabs}

\usepackage[pagebackref,breaklinks,colorlinks]{hyperref}

\usepackage{tikz}
\usepackage{comment}
\usepackage{amsmath,amssymb,bm} % define this before the line numbering.
\usepackage{tabularx}
\usepackage{multirow}
\usepackage{booktabs}
\usepackage{xcolor}

\usepackage{threeparttable}
\usepackage{adjustbox}
\usepackage{eqparbox}
\usepackage{arydshln}
\usepackage{wrapfig}
\usepackage{interval}

\usepackage{algorithm}
\usepackage{algpseudocode}

\makeatletter
\newcommand{\multiline}[1]{%
  \begin{tabularx}{\dimexpr\linewidth-\ALG@thistlm}[t]{@{}X@{}}
    #1
  \end{tabularx}
}
\makeatother

\newcommand{\distas}[1]{\mathbin{\overset{#1}{\kern\z@\sim}}}%
\newsavebox{\mybox}\newsavebox{\mysim}
\newcommand{\distras}[1]{%
	\savebox{\mybox}{\hbox{\kern3pt$\scriptstyle#1$\kern3pt}}%
	\savebox{\mysim}{\hbox{$\sim$}}%
	\mathbin{\overset{#1}{\kern\z@\resizebox{\wd\mybox}{\ht\mysim}{$\sim$}}}%
}

\title{3DArticCyclists: Generating Synthetic Articulated 8D Pose-Controllable Cyclist Data for Computer Vision Applications}

\author{Eduardo R. Corral-Soto, Yang Liu, Tongtong Cao, Yuan Ren, Liu Bingbing % <-this % stops a space
	\thanks{All authors are with Huawei Noah's Ark Lab, Canada at the time of writing.  
		{\tt\small \{eduardo.corral.soto, yang.liu9, caotongtong, yuan.ren3, liu.bingbing  \}@huawei.com}}%
}

\begin{document}

\maketitle
\thispagestyle{empty}
\pagestyle{empty}

%%%%%%%%%%%%%%%%%%%%%%%%%%%%%%%%%%%%%%%%%%%%%%%%%%%%%%%%%%%%%%%%%%%%%%%%%%%%%%%%
\begin{abstract}
In Autonomous Driving (AD) Perception, cyclists are considered safety-critical scene objects. Commonly used publicly-available AD datasets typically contain large amounts of car and vehicle object instances but a low number of cyclist instances, usually with limited appearance and pose diversity. This cyclist training data scarcity problem not only limits the generalization of deep-learning perception models for cyclist semantic segmentation, pose estimation, and cyclist crossing intention prediction, but also limits research on new cyclist-related tasks such as fine-grained cyclist pose estimation and spatio-temporal analysis under complex interactions between humans and
articulated objects. To address this data scarcity problem, in this paper we propose a framework to generate synthetic dynamic 3D cyclist data assets that can be used to generate training data for different tasks. In our framework, we designed a methodology for creating a new part-based multi-view articulated synthetic 3D bicycle dataset that we call 3DArticBikes that we use to train a 3D Gaussian Splatting (3DGS)-based reconstruction and image rendering method. We then propose a parametric bicycle 3DGS composition model to assemble 8-DoF pose-controllable 3D bicycles. Finally, using dynamic information from cyclist videos, we build a complete synthetic dynamic 3D cyclist (rider pedaling a bicycle) by re-posing a selectable synthetic 3D person, while automatically placing the rider onto one of our new articulated 3D bicycles using a proposed 3D Keypoint optimization-based Inverse Kinematics pose refinement. We present both, qualitative and quantitative results where we compare our generated cyclists against those from a recent stable diffusion-based method. 
\end{abstract}

%%%%%%%%%%%%%%%%%%%%%%%%%%%%%%%%%%%%%%%%%%%%%%%%%%%%%%%%%%%%%%%%%%%%%%%%%%%%%%%%
\section{Introduction}
\label{sec:intro}

In the context of Autonomous Driving (AD) Perception, cyclists are considered safety-critical scene objects. Deep learning-based computer vision models for tasks such as cyclist detection, segmentation, pose estimation and cyclist crossing intention prediction require large amounts of data for training. Commonly-used publicly-available AD datasets such as KITTI ~\cite{geiger2013vision}, NuScenes ~\cite{caesar2020nuscenes}, and Waymo ~\cite{sun2020scalability} typically contain large amounts of car/vehicle object instances but a relatively low number of cyclist instances usually with very limited appearance and pose diversity. This data scarcity problem is also present in Human-object interaction (HOI) applications in Embodied Artificial Intelligence (EAI) research, where there are insufficient labeled human-scene object pairs on the input images, and limited interaction complexity and granularity between them. Unlike cars and vehicles, which are \emph{rigid}, and can be treated as single-part objects, cyclists are a composition of a rider interacting dynamically with an articulated object (i.e. bicycle) which consists of multiple movable, usually rigid parts that are linked together by joints with kinematic constraints.
To overcome the human-articulated object data scarcity problem, recent methods have exploited Visual Language Models (VLM) ~\cite{zheng2023open,park2023viplo} and CLIP ~\cite{wang2024bilateral} to obtain semantic textual interaction descriptions from images that contain humans interacting with objects. Other methods address the data scarcity problem by creating new datasets. NeuralDome ~\cite{zhang2023neuraldome} introduced a multi-view dataset of humans interacting with $23$ rigid objects such as chairs, tables, and cooking tools. Instant-NVR ~\cite{jiang2023instant} uses monocular RGBD cameras and human/object tracking and key frame selection, coupled with neural radial fields to reconstruct and render interactions between humans and \emph{rigid} objects such as toys, backpacks, and a ``balance car''. I'm HOI ~\cite{zhao2024m} uses a combination of RGB camera with object-mounted IMU to capture 3D motions of both the human and rigid objects such as skateboard, gym weights, baseball bats, frying pans and umbrellas. PHYSCENE ~\cite{yang2024physcene} presents a method to synthesize physically-plausible dynamic interaction scenes for Embodied AI, but focuses on objects moving within a scene rather than interactions at the human body-articulated object level. The Align Your Gaussians ~\cite{ling2024align} is based on 3D Gaussians coupled with text-to-video Stable Diffusion models to animate synthetic objects suitable for games and movies, and presents interactions of characters with rigid objects such as guitars and swords. All these methods can generate interactions between humans/characters and rigid objects. However, more complex interactions between humans and articulated objects (e.g. a human rider pedaling a bicycle), suitable for training perception models have been left under-explored. 
\begin{figure} 
	\centering	
		\includegraphics[width=1.0\columnwidth, trim={0cm 15.5cm 0cm 0cm},clip]{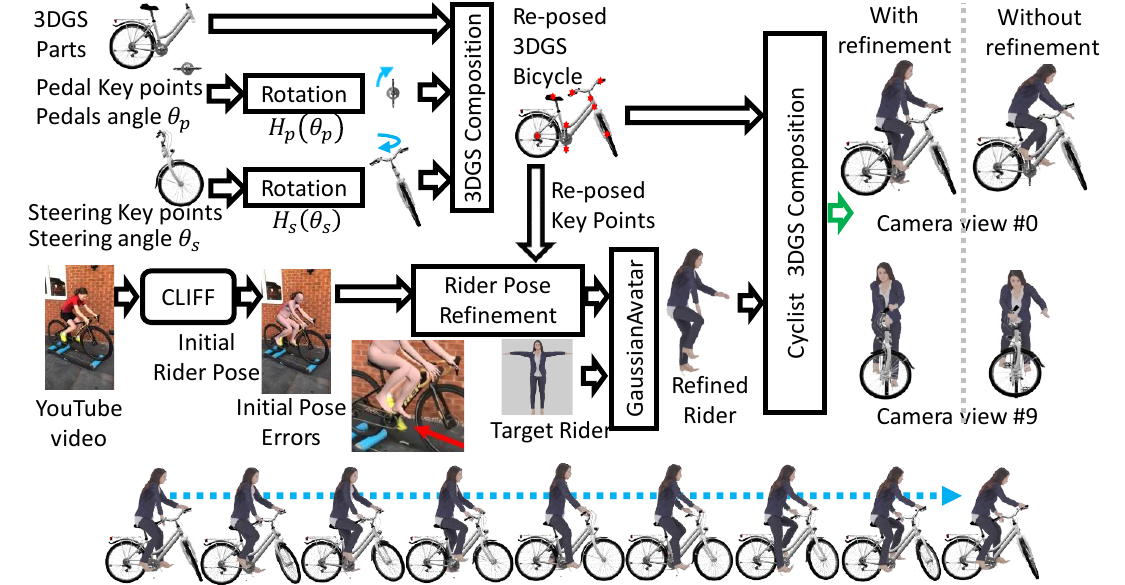} 		
	\caption{Proposed Pipeline to Generate Dynamic Synthetic 3D Cyclists.
	\label{top_cover_diagram} }
\end{figure}
%% trim={<left> <lower> <right> <upper>}

\textbf{Our contributions:} To address the 3D cyclist data scarcity problem, in this paper we propose a method to generate simulated 3D dynamic cyclist objects and interactions: \textbf{1)} We propose a methodology for creating a new part-based multi-view articulated synthetic 3D bicycle dataset that we call \emph{3DArticBikes}, which can be used to train 3D Gaussian Splatting (3DGS)-based 3D reconstruction methods. \textbf{2)} We propose a 3DGS-based parametric bicycle composition model to assemble pose-controllable (8-DoF) 3D bicycles (without a rider). \textbf{3)} We generate complete synthetic animatable dynamic 3D cyclists (rider pedaling a bicycle) by re-posing a selectable 3D person following the dynamics of a cyclist video, while automatically placing the rider onto one of our new articulated bicycles using a proposed 3D Keypoint optimization-based Inverse Kinematics pose refinement. We present both, qualitative and quantitative results where we compare our generated cyclists against those from a recent stable diffusion-based method. 

\section{Related Work}
\label{sec:related_work}
\textbf{3D Reconstruction and Rendering of Novel Views:} Neural Radiance Fields ~\cite{mildenhall2021nerf} (NeRF)-based 3D reconstruction methods can be roughly classified into \emph{Multi-view}, which require either multiple synchronized real camera, motion capture (MoCap) images or synthetic 3D data, and \emph{Monocular}, which operate on images captured by a single camera. The multi-view approaches can be further divided into methods that reconstruct \emph{stationary Scenes} (e.g. buildings, background) ~\cite{mildenhall2021nerf}, Nerfing-MVS ~\cite{wei2021nerfingmvs}, and methods that reconstruct \emph{stationary rigid foreground objects:} Neus ~\cite{wang2021neus}, and S-NeRF ~\cite{xie2023s}. NeRF-from-image ~\cite{pavllo2023shape}, and Zero-1-to-3 ~\cite{liu2023zero} are trained using multiple views rendered from rigid 3D objects.  3D Gaussian Splatting (3DGS) ~\cite{kerbl20233d}-based methods have gained popularity recently mainly because of their faster training speed and their reasonably high reconstruction and rendering quality of \emph{rigid} objects and \emph{static} scenes. Some notable NeRF-based methods for dynamic, deformable 3D human body reconstruction are Pifu ~\cite{saito2019pifu}, NeuMan ~\cite{jiang2022neuman}, HumaNeRF~\cite{weng2022humannerf}, and SHERF ~\cite{hu2023sherf}. These methods use the Skinned Multi-Person Linear model (SMPL) ~\cite{loper2023smpl} for human body representation and pose manipulation. HuGS ~\cite{moreau2024human} and GaussianAvatar ~\cite{hu2024gaussianavatar} are recent 3DGS and SMPL-based methods for 3D human body reconstruction and rendering. Typical datasets used to train these methods include THuman ~\cite{yu2021function4d} and  ZJU MoCap ~\cite{peng2021neural} (real images captured in the lab), and RenderPeople ~\cite{Renderpeople} (multi-view images of \emph{synthetic} 3D persons).

\textbf{Bicycle Reconstruction:}
The BIKED method ~\cite{regenwetter2022biked} represents bicycle features using PCA from bicycle models obtained from the BikeCAD website ~\cite{BIKECAD}. BIKED can generate novel 2D bicycle shapes, but does not generate 3D bicycles. ShapeNet ~\cite{chang2015shapenet} contains around $60$ non-realistic un-textured 3D bicycles. CO3D ~\cite{reizenstein2021common} and GSO ~\cite{downs2022google} include 3D models of scanned real rigid objects. The reconstructed quality of GSO's 3D bicycles and motorcycles is insufficient for our project. Both, the AKB-48 ~\cite{liu2022akb} and Gapartnet ~\cite{geng2023gapartnet} datasets include articulated 3D objects of indoor home and office object categories, such as Scissors, Eyeglasses, Book, Bottle, but do not include bicycles. A limited number of commercial articulated dynamic cyclist 3D models are available for purchase from websites like ~\cite{3DPeople}, which come with specific pre-selected (and fixed) rider and bicycle models. %As of the time of writing, we are not aware of the existence of any real-data multi-view motion-capture (MoCap) dataset for 3D dynamic bicycles or cyclists.
%%%%%%%%%%%%%%%%%%%%%%%%
%%%%%%%%%%%%%%%%%%%%%%%%

\section{Method}
\label{sec:method}
%%%%%%%%% BODY TEXT
We generate our own dataset of animatable/controllable synthetic 3D cyclists, where we split a cyclist into two main parts: 1) The rider part, and the rideable part, which can be a bicycle, a scooter or a motorcycle. For the rider part, we adopt the existing synthetic human body dataset from RenderPeople ~\cite{Renderpeople}, ~\cite{saito2019pifu}, ~\cite{hu2023sherf} which includes 360 Deg. views and camera matrices of 482 persons and is suitable for training recent 3D human body reconstruction methods.
\begin{figure*} 
	\centering	
		\includegraphics[width=1.7\columnwidth, trim={0cm 15.0cm 0cm 0cm},clip]{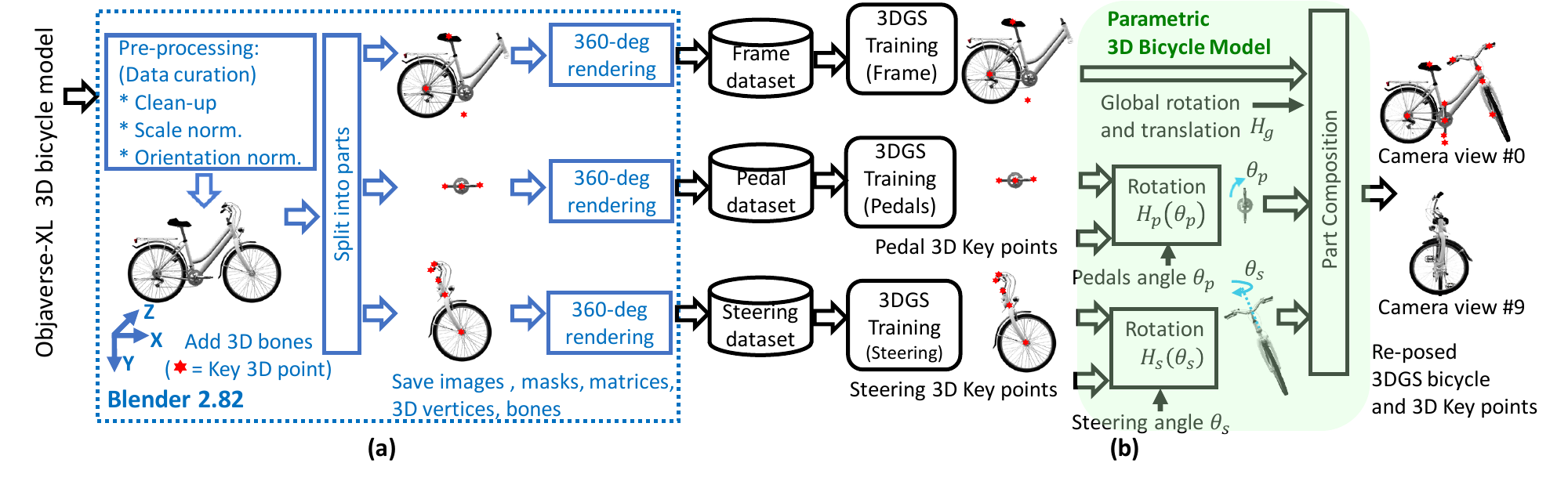} 		
	\caption{(a) Generating Raw 3D bicycle parts dataset in Blender for our articulated dynamic 3D cyclist generation project. (b) 3DGS bicycle steering and pedal angle manipulation. The bicycle steering and pedals Gaussian attributes are rotated by the desired angles $\theta_s$ and $\theta_p$ respectively, using $4 \times 4$ 3D rotation matrices. We then compose the outputted re-posed 3DGS bicycle by concatenating the parts Gaussians.
	\label{Combined_blender_data_generation_and_3dgs_parts_manipulation} }
\end{figure*}
%% trim={<left> <lower> <right> <upper>}

\section{Articulated 3DArticBikes Dataset}
%\section{Articulated Dynamic Cyclist Dataset}
\label{sec:bikedataset}
For the 3D bicycle part we designed a methodology for generating a new dataset that we called \emph{3DArticBikes}. Our starting point is the existing Objaverse-XL ~\cite{deitke2024objaverse} 3D object dataset which contains 3D rigid models of thousands of objects such as cars, toys, hydrants and bicycles. We extracted $23$ usable high-quality 3D rigid bicycle models. We discarded the rest since either they have non-realistic shapes, contain 3D clutter, or come without colored texture.

\textbf{3D Bicycle Data Pre-Processing (Curation).} We load each of the $23$ selected 3D bicycle models in Blender v2.82 ~\cite{Blender}, and apply the following pre-processing: 1) Remove non-bicycle objects, 2) Normalize the scale, 3) Translate and rotate the bicycles to a canonical 3D location/orientation standing up on the ground $X-Z$ plane, with the pedals axle centroid directly above the 3D origin, and the front wheel looking along the X-axis (see 3D coordinate system in Fig. \ref{Combined_blender_data_generation_and_3dgs_parts_manipulation}(a)). We articulate the originally-rigid bicycles by splitting the bicycle model (3D vertex groups ~\cite{Blender}) into three separate parts: 1) Frame and rear wheel,  2) Pedals, and 3) Steering and front wheel (see Fig. \ref{Combined_blender_data_generation_and_3dgs_parts_manipulation}(a)).  %Note: 

\textbf{Articulated bicycle parts and 3D key points.} We introduced armature bones ~\cite{Blender} to be able to rotate the steering and the pedals about their axes in Blender. We introduced bones for the following bicycle parts:  1) Seat (saddle),   2) Steering axle, 3) Left handle, 4) Right handle, 5) Pedals axle, 6) Left pedal, 7) Right pedal, 8) Front and rear wheel rotation axles, and 9) Ground/origin. We saved the 3D bone and 3D keypoints (see red markers in Fig. \ref{Combined_blender_data_generation_and_3dgs_parts_manipulation}(a)).

\textbf{360-Deg. Views Rendering in Blender.} We follow the RenderPeople dataset folder structure and format used in ~\cite{hu2023sherf} to be able to train NeRF and 3DGS-based 3D reconstruction models with our new 3D bicycle parts datasets. We define a camera $P$ at a canonical 3D pose with rotation matrix  $R=I$ and camera location $C=[0, 0, -12]$ m, where $I$ is the $3 \times 3$ identity matrix. We adopt the same focal length $fx=fy=2084.97$ used in ~\cite{hu2023sherf}. We then rotate the camera $P$ around each of the three 3D bicycle parts at $36$ discrete azimuth angular positions ($36$ camera views) with increments of 10 Deg., capturing a $512 \times 512$ RGBA image, including a binary mask at each camera pose, and save them together with the 3D bones/Keypoints information, the intrinsic/extrinsic camera matrices, and the bicycle part 3D points - see Fig. \ref{Combined_blender_data_generation_and_3dgs_parts_manipulation}(a).

%%%%%%%%%%%%%%%%%%%%%%%%
%%%%%%%%%%%%%%%%%%%%%%%%
\section{Building Animated 3DGS Cyclists.}
\label{sec:3dgs_cyclist_parts_anim_composition}
We now have multi-view 3D datasets for training 1) a model for the rideable object (using the 3DArticBikes dataset), and 2) a rider model (using the RenderPeople dataset). 3DGS has recently become a popular method for representing and reconstructing 3D objects because of its short training time, portability/deployability,  its versatility and ease of manipulation (rotation and translation) of 3D Gaussians, and the ease of integration of 3DGS foreground objects into existing 3DGS background scenes, which makes it a preferable representation compared to NeRF or to working directly with graphics packages like Blender.  Therefore, we adopt it for 3D object representation. For the rider part, we train GaussianAvatar ~\cite{hu2024gaussianavatar}, which learns a human body Gaussian representation that can be re-posed using a target body pose in SMPL format (see Fig. \ref{top_cover_diagram}). For the rideable (Bicycle) part, we train the vanilla 3DGS from ~\cite{kerbl20233d} to learn a Gaussian representation for each of the \emph{ three separate} rigid bicycle parts, each in canonical pose. 

%\subsection{3DGS Rideable (Bicycle) Articulation.}
\subsection{3DGS-Based Parametric Articulated Bicycle Model.}
\label{sec:bike_articulation}
We define the 8-DoF pose of a 3D bicycle as follows: $pose_{3D} = \{ \theta_p, \theta_s, \theta_{X}, \theta_{Y}, \theta_{Z},  t_X, t_Y, t_Z    \}$, where, $\theta_p$ and $\theta_s$ are the bicycle pedals axle and steering angles about their rotation shafts on the bicycle body frame, $\theta_{X}$, $\theta_{Y}$, and $\theta_{Z}$ are the bicycle body frame rotation angles about the world's $X$, $Y$, and $Z$ 3D axes respectively, and $(t_X$, $t_Y$, $t_Z)$ define the bicycle body 3D translation from the 3D origin. Once a 3DGS model has been trained for each of the rigid bicycle parts, we designed a parametric 3D bicycle model (see Fig. \ref{Combined_blender_data_generation_and_3dgs_parts_manipulation}(b)) to rotate, in inference mode, the Gaussian attributes (3D points, rotations, scales, and spherical harmonics) ~\cite{kerbl20233d} of both, the bicycle steering and pedals by the desired angles $\theta_s$ and $\theta_p$ using the $4 \times 4$ SE(3) 3D rotation matrices $H_{s}(\theta_s)$ and $H_{p}(\theta_p)$ respectively. We define $H_{s}(\theta_s)$ as:
\begin{equation} 
\begin{aligned} 
\label{eq:Hs}
H_{s} =  H_{sz}(\theta_{vy, v}, k_{s1}, k_{s2}) H_{sx}(\theta_s)    H_{sz}^{-1}(\theta_{vy}, k_{s1}, k_{s2}),
\end{aligned}
\end{equation}
where $\theta_{vy}$ is the angle on the $X-Y$ plane of a unit vector $v=(v_x, v_y, v_z)^T$ defined by two Keypoints $k_{s1}, k_{s2}$ along the steering shaft, and $H_{sz} \in SE(3)$ rotates 3D points about the $Z$ 3D axis, and  $H_{sx} \in SE(3)$ rotates 3D points about the $X$ 3D axis by $\theta_s$. We define $H_{p}(\theta_p)$ as:
\begin{equation} 
\begin{aligned} 
\label{eq:Hp}
H_{p} =  H_{pc}(v_p) H_{pz}(\theta_p)    H_{pc}^{-1}(v_p),
\end{aligned}
\end{equation}	
where $v_p$ is the 3D location of the pedal shaft centroid, $H_{pc} \in SE(3)$ translates 3D points to the origin, and $H_{pz} \in SE(3)$ rotates 3D points about the $Z$ 3D axis by $\theta_p$. We also apply the same transformations to the bicycle 3D Keypoints, since these will be used in the refinement of the rider's body pose at a later stage in our pipeline. We then assemble/compose a complete re-posed 3DGS bicycle by concatenating the reposed 3DGS parts with the bicycle frame. Fig. \ref{Combined_blender_data_generation_and_3dgs_parts_manipulation}(b) shows an example of steering and pedal angle manipulation in Gaussian space for $\theta_s=90 Deg$ and $\theta_p=90 Deg$.

\subsection{3DGS Rider (Human) Animation.}
\label{sec:rider_animation}
There are many SMPL-based methods to animate a 3D human body using videos ~\cite{sun2021monocular}, ~\cite{kolotouros2019learning}, ~\cite{jiang2022neuman}, ~\cite{shen2023x}. In our work we animate the 3D rider using dynamic information from publicly-available YouTube cyclist videos by inputting the images into the CLIFF model ~\cite{li2022cliff}, which enables us to obtain a noisy estimate of the rider's body shape and pose in SMPL format, and estimates of the global 3D location and orientation  $H_{g}(\theta_{X}, \theta_{Y}, \theta_{Z},  t_X, t_Y, t_Z)$ ~\cite{li2022cliff}, ~\cite{kolotouros2019learning} of the rider's body for each incoming video image, where $H_{g} \in SE(3)$.

\subsection{Placing the Rider on the Bicycle}
\label{sec:rider_pose_refinement}
\begin{figure} 
	\centering	
		\includegraphics[width=0.45\columnwidth, trim={5cm 18.5cm 5cm 0cm},clip]{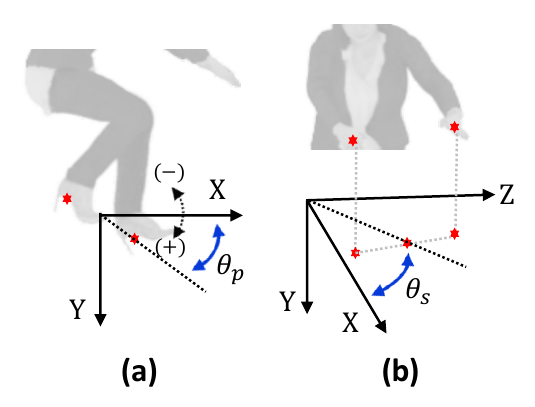} 		
	\caption{(a) Deriving the left pedal angle from the rider's ankle 3D Keypoints. (b) Deriving the steering angle from the rider's wrist 3D Keypoints.
	\label{Deriving_pedal_and_steering_angles_from_cliff} }
\end{figure}

The bicycle pose parameters $\theta_s$, $\theta_p$, and $H_g$ can be set: 1) Manually or programmatically. For example, in Fig. \ref{top_cover_diagram} (bottom) the values of $\theta_s$, and $\theta_p$ are being varied linearly from 
$-90$ to $90$ Deg (steering), and from $-180$ to $180$ Deg (pedals).  2) $\theta_s$ and $\theta_p$ can be derived using geometry from the rider's wrists and ankle joints estimated by CLIFF, and 3) A 3D bicycle pose estimation model can be designed to estimate them directly from the input image, which is out of the scope of this paper.
%. However, CLIFF errors propagate into the 3D bicycle re-posing, which is undesirable. 
For cyclist animation from videos we derive the values of $\theta_s$, $\theta_p$, and $H_g$ from the estimated rider's pose provided by CLIFF as explained next. We rotate the rider's body 3D joints to canonical bicycle pose. We then use simple geometry (i.e. arctan) and the orthographic projections of the rider's ankle 3D keypoints onto the $X-Y$ plane to derive $\theta_p$ (see Fig. \ref{Deriving_pedal_and_steering_angles_from_cliff}). Similarly, we use the mid-point along a line that crosses through the projections of the rider's wrists onto the $X-Z$ plane to derive $\theta_s$.
%% trim={<left> <lower> <right> <upper>}

\textbf{Rider-Rideable 3D Keypoint-Based Pose Refinement.} In the SMPL human body model ~\cite{loper2023smpl}, the pose is represented by a set $\theta=\{  \mathbf{ \theta_k }  \}, k=[1,2,...,K]$ of $K=24$ local 3D joint rotations represented as axis-angle 3-vectors $\mathbf{\theta_k}$. These joint rotations correspond to human body joints such as "Pelvis", "Left hip", "Left knee",  "Left ankle", "Right shoulder", "Right elbow", "Right wrist", etc. Given $\theta$, and the body shape parameters $\beta \in \mathbb{R}^{10}$ ~\cite{loper2023smpl}, the SMPL model $\mathcal{M}=(\theta, \beta)$ returns the body mesh $M \in \mathbb{R}^{N \times 3}$, with $N=6890$ 3D vertices.   
And the set of 3D body joints $J=\{  \mathbf{ J_k }  \}$ is obtained as a linear combination of the vertices using a trained regressor ~\cite{loper2023smpl}.
We input an image $I$ of a cyclist to CLIFF, to obtain estimated rider's body parameters $(\theta_{c}, \beta_{c})$. We observe that the body pose $\theta_{c}$ is reasonably accurate, but it often comes with errors at specific body locations, such as 3D location mis-match between the rider's hands and the bicycle handles, and between the rider's feet and bicycle pedals (see left side of Fig. \ref{top_cover_diagram}).
To address this problem we propose an iterative Inverse Kinematics (IK) optimization-based 3D human body pose refinement that leverages the controllable 3D pose from the bicycle to align the 3D locations of the problematic human body joints such as wrists, pelvis, and ankles to the corresponding bicycle 3D Keypoints (i.e. steering handles, seat, and pedals).
Given the parameters $(\theta_{c}, \beta_{c})$ and the bicycle 3D Keypoints (reposed using $\theta_s$, $\theta_p$, $H_{g}$), we split the human pose parameters as $\theta_{c} = \theta_{c_{fixed}} \cup \theta_{c_{rfn}} $. Where,
\begin{equation} 
\begin{aligned} 
\label{eq:cliff_pose}
\theta_{c_{rfn}} = \{  &  \bm{ \theta_{bbtn} } ,\bm{ \theta_{Lsho}} , \bm{ \theta_{Rsho}} , 
                                      \bm{ \theta_{Lelb}} , \bm{ \theta_{Relb}} , \bm{ \theta_{Lhip}} , \\
                                 &  \bm{ \theta_{Rhip}} , \bm{ \theta_{Lknee}} , 
                                     \bm{ \theta_{Rknee}} , \bm{ \theta_{Lank}} , \bm{ \theta_{Rank}}     
                             \},
\end{aligned}
\end{equation}
and the subscripts ``bbtn'', ``sho'', ``elb'', ``hip'', ``knee'', and ``ank'' denote ``belly button'', ``shoulder'', ``elbow'', ``hip'', ``knee'', and ``ankle'' which represent 3D joint rotations , and ``L', ``R'' denote ``Left'' and ``Right'' human body sides.  And $\theta_{c_{fixed}} $ are the rest of the body joint rotations, which we keep fixed (unchanged) during the optimization process.
We then define a body pose residual $\Delta_{\theta_{rfn}}$ as:
\begin{equation} 
\begin{aligned} 
\label{eq:delta_pose}
\Delta_{\theta_{rfn}} = \{  & \bm{ \Delta_{\theta_{bbtn}} },\bm{ \Delta_{\theta_{Lsho}} }, \bm{ \Delta_{\theta_{Rsho}} }, 
                                          \bm{ \Delta_{\theta_{Lelb}} }, \bm{ \Delta_{\theta_{Relb}} }, \bm{ \Delta_{\theta_{Lhip}} }, \\
                                      & \bm{ \Delta_{\theta_{Rhip}} }, \bm{ \Delta_{\theta_{Lknee}} }, 
                                         \bm{ \Delta_{\theta_{Rknee}} }, \bm{ \Delta_{\theta_{Lank}} }, \bm{ \Delta_{\theta_{Rank}} }    
                                  \}.
\end{aligned}
\end{equation}
We implement an iterative optimization method that updates $\theta_{c_{rfn}}(i+1) = \theta_{c_{rfn}}(i) +  \Delta_{\theta_{rfn}}(i)$, where $i$ denotes the iteration index. At each iteration we minimize the following Chamfer Distance objective function:
\begin{equation} 
\begin{aligned} 
\label{eq:loss_pose_refinement}
\mathcal{L}_{rfn} = \mathcal{CH}( & \{ x_{Lwrist}, x_{Rwrist}, x_{pelvis}, x_{Lank}, x_{Rank} \} , \\
                                                  & \{ x_{Lhandle}, x_{Rhandle}, x_{seat}, x_{Lped}, x_{Rped}  \} ),                               
\end{aligned}
\end{equation}
which encourages the 3D spatial aligment of the human body 3D Keypoints $x_{Lwrist}$, $x_{Rwrist}$, $x_{pelvis}$, $x_{Lank}$, $x_{Rank}$ to the bicycle 3D Keypoints $x_{Lhandle}$, $x_{Rhandle}$, $x_{seat}$, $x_{Lped}$, $x_{Rped}$ respectively. In summary, the optimization process places the rider's hands, ankles and pelvis near the bicycle handles, pedals, and seat respectively. We use the Adam optimizer with LR=$0.05$, and we set (experimentally) the number of iterations to $50$. We select the $\hat{\Delta}_{\theta_{rfn}}(i)$ residuals that minimize Eqn. (\ref{eq:loss_pose_refinement}). Once the optimization procedure ends, we use the refined rider's body pose $\theta_{copt} = \theta_{c_{fixed}} \cup   (\theta_{c_{rfn}} + \hat{\Delta}_{\theta_{rfn}}(i) ) $ in the input parameters $(\theta_{copt}, \beta_{c})$ to the the GaussianAvatar model to obtain the final refined rider's 3D body in 3DGS format given a selected RenderPeople person (see Fig. \ref{top_cover_diagram}). Finally, we assemble (in canonical pose) the complete 3D cyclist by concatenating the 3D Gaussians from the re-posed 3D bicycle together with the \emph{refined} re-posed rider's body, and apply the global 3D location and rotation $H_g$ estimated by CLIFF. We are now ready to render the controllable 3D cyclist at a desired 8-DoF pose and camera view, given a selected video frame. The composition mechanism we just described enables us to select different rider-bike combinations: There are $482$ riders from RenderPeople, and $23$ 3D bicycles from 3DArticBikes, which gives us $11,086$ possible rider-bicycle combinations to generate different 3D cyclists. From this point on, we refer to our generated 3D cyclists as \emph{3DArticCyclists}.
\begin{figure*} 
	\centering	
		\includegraphics[width=1.9\columnwidth, trim={0cm 9.0cm 0cm 0cm},clip]{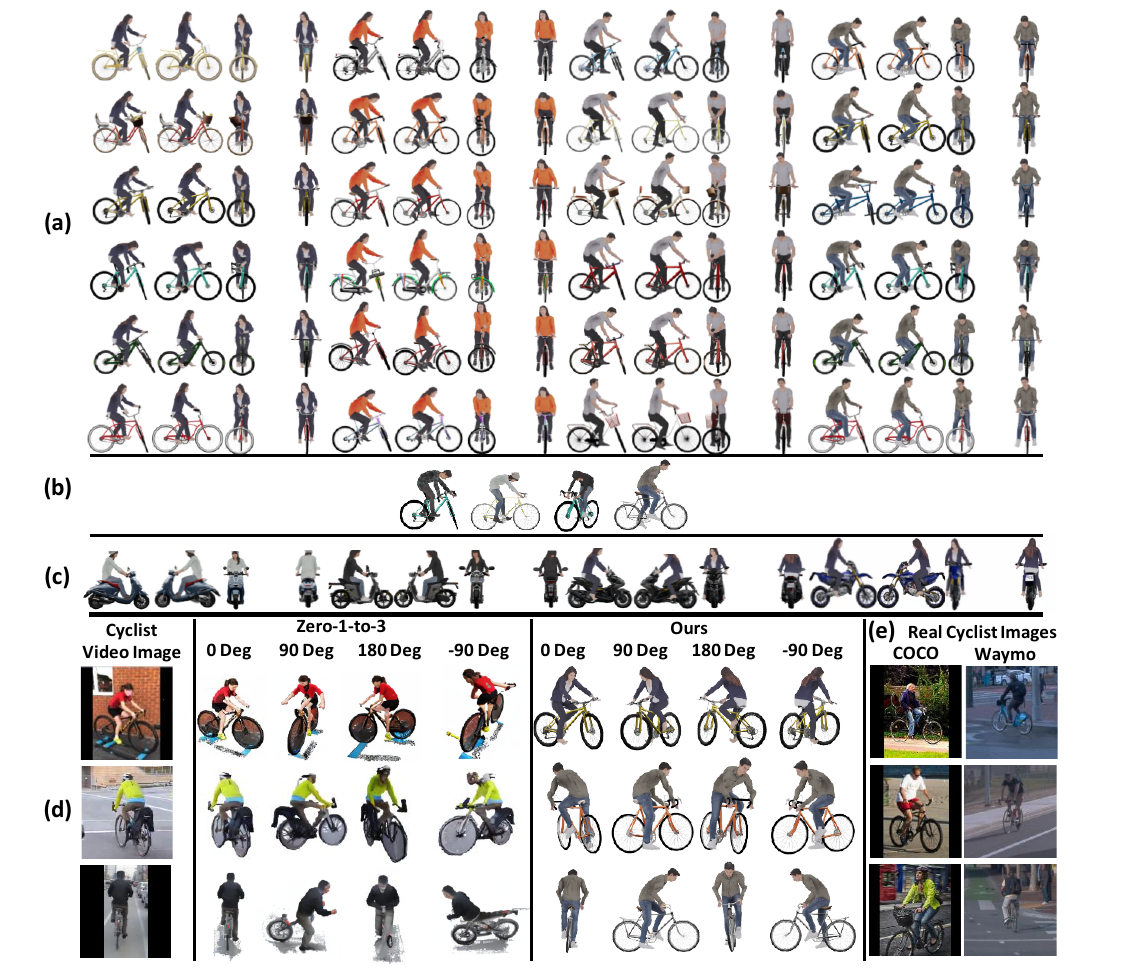} 		
	\caption{Example qualitative results. (a) Different 3D persons from RenderPeople riding bicycles from our 3DArticBikes dataset, (b) Typical failure cases: Seat too high for some riders, gaze/head pose looking down, rider not properly seated. (c) We tested the generalization our pipeline with scooters and motorcycles. (d) Cyclist generation with animation from cyclist videos. We compared our method against the Zero-1-to-3 method. (e) Real COCO and Waymo images used to compute FID/KID metrics from Table \ref{eval_table_fid_kid}.
	\label{qualitative_results_fig} }
\end{figure*}
%% trim={<left> <lower> <right> <upper>}

\section{Experiments and results}
\label{sec:experiments}
\subsection{Qualitative Results}
\label{sec:qualitative}
\textbf{Controlling $\theta_{s}$ and $\theta_{p}$ Manually, with Different Rider/Bicycle Combinations.} 
We demonstrate the versatility of our 3DArticCyclist pipeline by generating
articulated cyclists of different rider-bicycle combinations with two manually-set values of the pedal and steering angles $\theta_p$ and $\theta_s$. Fig. \ref{qualitative_results_fig} (a) shows four different riders on different 3DArticBike bicycles under two camera views: $0$ and $90$ Deg. with two angular values for $\theta_{s}$ and $\theta_{p}$. Fig. \ref{qualitative_results_fig} (c): We tested the generalization of our pipeline by placing riders on 3D scooters and motorcycles downloaded from ~\cite{3Dmodels}. Scooters and motorcycles are easier to process than bicycles due to the removal of the pedal rotation $\theta_{p}$ degrees of freedom. 

\textbf{Controlling $\theta_{s}$ and $\theta_{p}$ Using the Dynamics From a Cyclist Video.} 
We generated $200$-frame sequences of animated 3D riders on bicycles from our 3DArticBike dataset. We used the dynamic information from three publicly-available YouTube videos: Riding on Rollers ~\cite{YouTube_Rollers}, Flying By Traffic ~\cite{YouTube_Flying}, and Beijing Commute ~\cite{YouTube_Beijing}. In this setting we derive the pedal and steering angles $\theta_p$ and $\theta_s$ from the re-posed rider body as explained in Section \ref{sec:rider_pose_refinement}. We compare our method against Zero-1-to-3 ~\cite{liu2023zero} which is a generative method based on stable-diffusion capable of manipulating the camera view and reconstructing and rendering 3D objects given an input image. Fig. \ref{qualitative_results_fig} (d) shows generated cyclists from both our method and Zero-1-to-3. Note that Zero-1-to-3 attempts to reconstruct and preserve the appearance from the input image cyclist. However, any change of camera (or cyclist) pose results in large 3D cyclist shape and appearance distortions. Our method does not attempt to reconstruct/preserve appearance, but it focuses on preserving the pose of the rider from the input image while generating a 3D bicycle 8-DoF pose that is consistent with the rider's pose, hence generating higher-quality practical/usable animatable 3D cyclists.
\begin{table}
	\caption{Perceptual quality scores computed with real cyclist images from the COCO 2017 and Waymo datasets.} 
	\label{eval_table_fid_kid}
	\centering
	\begin{tabular}{lll}
		\toprule
		%\cmidrule(r){1-2}		
		3D Cyclist                                       & COCO Cyclists                                       & Waymo Cyclists                                      \\
		Generation Method                          & FID $\downarrow$ KID $\downarrow$   & FID $\downarrow$ KID $\downarrow$         \\
		\midrule		
		Zero-1-to-3 ~\cite{liu2023zero}       & 175.0,   0.179                                       & 207.0,   0.230                                         \\		
		3DArticCyclists (\textbf{ours})            & \textbf{169.2},   \textbf{0.165}          & \textbf{181.0},   \textbf{0.190}                     \\
		\bottomrule
	\end{tabular}
\end{table}

\subsection{Quantitative Results}
\label{sec:quantitative}
\textbf{Perceptual Quality Scores.} We follow the evaluation methodology presented by recent human-scene interaction methods ~\cite{yang2024physcene}, ~\cite{li2024genzi} for the case when no ground-truth is available to evaluate the quality of their generated scenes. In our method, the generated 3D cyclists do not have 3D ground-truth either, because: 1) The rider's pose is an estimated quantity (CLIFF), and 2) Because our method alters (refines) the CLIFF rider's pose to follow the 3D Keypoints form our 3D bicycles with shapes and appearances that do not match those from the input image. Therefore, we follow PHYSCENE ~\cite{yang2024physcene}, and compute the Fr\'{e}chet Inception Distance (FID) ~\cite{heusel2017gans}, and Kernel Inception Distance (KID) ~\cite{binkowski2018demystifying} perceptual scores, which are based on a distance between features from real and generated images, where a low score is desirable. We used the same videos from Section \ref{sec:qualitative} to generate animated 3D Cyclists using our method, and rendered images for four camera azimuth views: $0$, $90$, $180$, and $-90$ Deg. For each video we generated $800$ output images. We followed a similar procedure to process the same three image sequences using Zero-1-to-3 to generate an output image for each of the four camera views. To compare features against real, natural images, we extracted cyclist images from two publicly-available datasets:  1) Waymo ~\cite{sun2020scalability}, and 2) COCO $2017$ ~\cite{lin2014microsoft} from which we selected images of non-truncated cyclists with a bounding box max. dimension greater than 60 pixels, and centered/padded/rescaled the sub-images to have the same dimensions as ours and Zero-1-to-3. We show examples of the Waymo and COCO images in Fig. \ref{qualitative_results_fig} (e). Table \ref{eval_table_fid_kid} shows the results.  Note that the magnitude of the FID and KID metrics seem relatively high compared to those from PHYSCENE. We believe that this happens because we are comparing features between real and synthesized images, and because PHYSCENE uses a scaling factor of $0.001$. Nevertheless, our FID/KID numbers are lower than those from Zero-1-to-3, which is consistent with the results observed in Fig. \ref{qualitative_results_fig} (d).   

\textbf{Using 3DArticBikes Data to Improve Bicycle Semantic Segmentation Performance.} As an application example, we use generated cyclist data from our framework to train the Mask2Former ~\cite{cheng2022masked} semantic segmentation model. One of the datasets used to train Mask2Former is COCO ~\cite{lin2014microsoft} for which the "cyclist" class does not exist. Therefore, we use bicycles from 3DArticBikes instead of cyclists. We generate $2500$ bicycle images with random 8-DoF poses ($\theta_{p}$, $\theta_{s}$, $H_{g}$), for each of our $23$ 3D bicycles and prepare training data following the COCO format. We then follow the next steps: 1) Resume training of a base pre-trained model provided by the authors on the COCO train2017 set, and evaluate on COCO val2017 set. 2) Augment the COCO train2017 set by adding our data. 3) Repeat the training from the first step, but now with the COCO train2017 set augmented with our bikes data, and evaluate on COCO val2017 set. Table \ref{eval_table_semanticsegm} shows that both of the bicycle and motorcycle classes improved by $1.02$ and $1.84$ respectively. We believe that motorcycles improvement is due to shape similarities to bicycles.

%\begin{table}
%	\caption{Data Augmentation Experiment With Mask2Former Semantic Segmentation Model.} 
%	\label{eval_table_semanticsegm}
%	\centering
%	\begin{tabular}{llll}
%		\toprule
%		%\cmidrule(r){1-2}		
%		                                                              & AP $\uparrow$   &                        &                          \\         
%		Experiment                                             & Overall               & Bicycle             &  Motorcycle        \\
%		\midrule		
%		Mask2Former ~\cite{cheng2022masked}  & 44.09                & 22.34               & 39.73                \\		
%		Mask2Former+3D Bikes                           & \textbf{44.33}   & \textbf{23.36}  & \textbf{41.57}                \\
%		\bottomrule
%
%	\end{tabular}
%\end{table}	

\begin{table}
	\caption{Data Augmentation Experiment With Mask2Former Semantic Segmentation Model.} 
	\label{eval_table_semanticsegm}
	\centering
	\begin{tabular}{llll}
		\toprule
		%\cmidrule(r){1-2}		
		                                                             & AP $\uparrow$   &                      & Overall         \\         
		Experiment                                             & Bicycle             &  Motorcycle      & (80 classes)   \\
		\midrule		
		Mask2Former ~\cite{cheng2022masked}   & 22.34               & 39.73                & 44.09          \\		
		Mask2Former+3D Bikes (\textbf{ours})        & \textbf{23.36}   & \textbf{41.57}   &  44.33          \\
		\bottomrule

	\end{tabular}
\end{table}

\subsection{Ablation studies}
\label{sec:ablation}
\begin{figure} 
	\centering	
		\includegraphics[width=0.7\columnwidth, trim={0cm 11.0cm 0cm 0cm},clip]{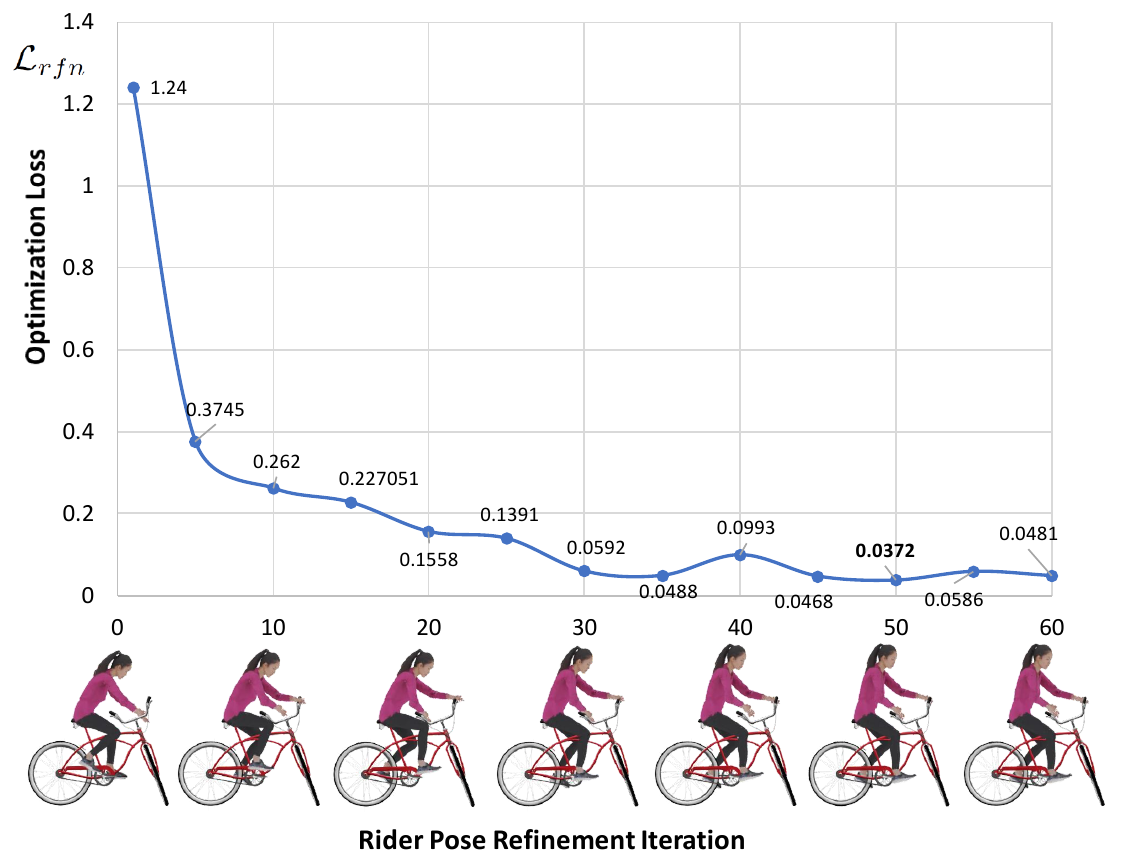} 		
	\caption{Rider Pose Refinement Inverse Kinematics Optimization Iterations.
	\label{IK_iterations_exper} }
\end{figure}

Fig. \ref{IK_iterations_exper} shows the results of running the pose refinement on a fixed rider-bicycle combination for different numbers of iterations. The plot explains our choice of setting the maximum number of iterations to $50$. The average run times to generate a 3D cyclist (excluding loading/saving data from/to files) are:  1) The reposing and assembling 3D Bicycles (without a rider) takes 0.32 Sec., 2) Each iteration of rider's body refinement (IK) takes 0.86 Sec. We use a maximum of 50 iterations. 3) The final rider's body reposing using GaussianAvatar, plus assembling the final outputted 3D cyclist (rider on the bicycle) takes 1.74 Sec.
%BLENDER TIME TO RENDER a single bicycle (no cyclist):  from 17 to 20 Sec per image.  
%% trim={<left> <lower> <right> <upper>}

\section{Limitations}
\label{sec:limitations}
Fig. \ref{qualitative_results_fig} (b) shows the typical failures that we observe in our generation pipeline: 1) Extreme rider's back position due to the seat being set too high (rider's feet cannot reach the pedals), 2) Riders not properly seated on the saddle. The appearance diversity is limited to the available 3D riders and 3D bicycle models. We have assumed that both rider's hands are on their corresponding bicycle handles (and that their feet are on their pedals). This assumption can be easily broken. Our parametric bicycle model ignores the individual pedal and wheel rotations. We have used the basic SMPL model. To improve the expressiveness of the rider's pose, the SMPL-X model could be used instead.

\section{Conclusions}
\label{sec:conclusions}
We have presented a method to generate 8-DoF pose-controllable 3D dynamic bicycles and cyclists that can be used in human-object interaction and autonomous driving perception applications. We have demonstrated through experimental evaluations that the perceptual quality of our generated cyclists is better that that of a state-of-the art stable-diffusion method, as well as the applicability of our cyclists to semantic segmentation performance improvement on bicycles. We hope that these new dynamic cyclists help enable new research on spatio-temporal analysis and pose estimation under complex interactions between humans and articulated objects that involve occlusions and motion.

%%%%%%%%%%%%%%%%%%%%%%%%%%%%%%%%%%%%%%%%%%%%%%%%%%%%%%%%%%%%%%%%%%%%%%%%%%%%%%%%

%%%%%%%%%% REFERENCES
\clearpage

\bibliographystyle{IEEEtran}
\bibliography{egbib_cyclist_arxiv_2025}
%%\bibliography{density_p2p_master_egbib_references}

\end{document}